\documentclass[11pt,a4paper]{article}
\usepackage[hyperref]{eacl2021}
\usepackage{times}
\usepackage{latexsym}

\usepackage{microtype}

\usepackage{graphicx}

\usepackage{caption}
\usepackage{subcaption}
\usepackage{amsfonts}
\usepackage{booktabs}
\usepackage{multirow}
\usepackage{xcolor}

\aclfinalcopy 

\title{Deep Multi-Task Model for Sarcasm Detection and Sentiment Analysis in Arabic Language}

\author{Abdelkader El Mahdaouy$^1$ \hspace{0.5cm} Abdellah El Mekki$^1$ \hspace{0.5cm} Nabil El Mamoun$^2$ \\\textbf{Kabil Essefar$^1$ \hspace{0.5cm} Ismail Berrada$^1$ \hspace{0.5cm} \textbf{Ahmed Khoumsi}$^3$}\\
$^1$School of Computer Sciences, Mohammed VI Polytechnic University, Morocco \\
$^2$Faculty of Sciences Dhar EL Mahraz, Sidi Mohamed Ben Abdellah University, Morocco\\
$^3$Dept. Electrical \& Computer Engineering, University of Sherbrooke, Canada\\
{\tt \{firstname.lastname\}@um6p.ma} \\
{\tt ahmed.khoumsi@usherbrooke.ca}\\
}

\date{}

\begin{document}
\maketitle
\begin{abstract}
The prominence of figurative language devices, such as sarcasm and irony, poses serious challenges for Arabic Sentiment Analysis (SA). While previous research works tackle SA and sarcasm detection separately, this paper introduces an end-to-end deep Multi-Task Learning (MTL) model, allowing knowledge interaction between the two tasks. Our MTL model's architecture consists of a Bidirectional Encoder Representation from Transformers (BERT) model, a multi-task attention interaction module, and two task classifiers. The overall obtained results show that our proposed model outperforms its single-task counterparts on both SA and sarcasm detection sub-tasks.  
\end{abstract}

\section{Introduction}
The popularity of the Internet and the unprecedented reach of social media platforms allow users to express their opinions on a wide range of topics. Thereby, Sentiment Analysis (SA) has become a cornerstone for many applications such as digital marketing, product review analysis, customer feedback, social media monitoring, etc. SA consists of determining the expressed sentiment (positive, negative, or neutral) conveyed by a text or a piece of text.

Over the past decade, significant research advances have been achieved for Arabic SA \cite{Badaro:2019Talip, ALAYYOUB2019320,OUESLATI2020408,ABUFARHA2021102438}. However, the mutual interaction and impact of figurative language devices, like sarcasm and irony, and Arabic SA remain under-explored \cite{farha2020arabic,ABUFARHA2021102438,AbbesZEA20}. These latter devices allow us to express ourselves intelligently beyond the literal meaning of words. Although the literature uses the terms irony and sarcasm interchangeably, they have different meanings and there is no consensus on their definition \cite{rosso2016,FARIAS2017113,ZHANG20191633}. Both sarcasm and irony devices pose a real challenge for SA as they can reverse the expressed sentiment polarity from positive to negative \cite{FARIAS2017113,farha2020arabic,ABUFARHA2021102438}. Therefore, there is an urgent need to develop sarcasm-aware SA tools. 

Previous research works on Arabic SA and sarcasm detection have dealt with both tasks separately \cite{Ghanem2019,GhanemKBRM20,AbbesZEA20,farha2020arabic}. \citet{AbbesZEA20} have built a corpus for irony and sarcasm detection in Arabic language from twitter using a set of ironic hashtags. Unlike the previous work of \citet{KAROUI2017161} that have relied on ironic hashtags to label the tweets, the annotation is performed manually by two Arabic language specialists. \citet{ABUFARHA2021102438} have presented an overview of existing Arabic SA methods and approaches, and a benchmarking using three existing datasets. Their results have shown that most of the evaluated models  perform poorly on the SemEval and ASTD datasets.  Due to the label inconsistencies  discovered, they have re-annotated the previously mentioned datasets for SA and sarcasm detection. In addition to the highly subjective nature of SA task, they have reported a large performance drop in the case of sarcastic tweets \cite{farha2020arabic,ABUFARHA2021102438}. 
\begin{figure*}[ht]
     \centering
     \label{fig:arSar}
     \begin{subfigure}[b]{0.3\textwidth}
         \centering
         \includegraphics[scale=0.36]{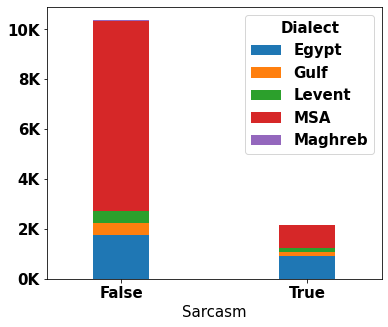}
         \caption{Distribution of sarcastic tweets per region}
         \label{fig:sar_reg}
     \end{subfigure}
     \hfill
     \begin{subfigure}[b]{0.3\textwidth}
         \centering
         \includegraphics[scale=0.36]{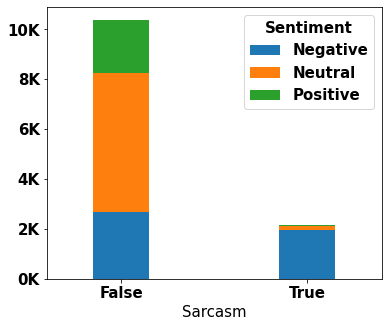}
         \caption{Distribution of sarcastic tweets per sentiment polarity}
         \label{fig:sar_sen}
     \end{subfigure}
    \hfill
     \begin{subfigure}[b]{0.3\textwidth}
         \centering
         \includegraphics[scale=0.36]{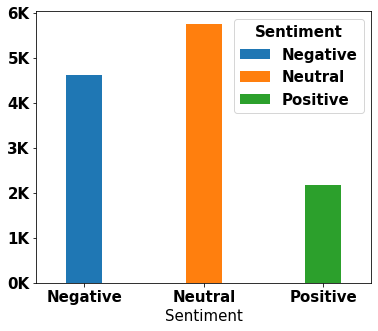}
         \caption{Distribution of sentiment polarities\\}
         \label{fig:sen_dist}
     \end{subfigure}
    \caption{ArSarcasm-v2 dataset: distribution of sarcastic tweets and their sentiment polarities. True and False denote sarcastic and non-sarcastic tweets, respectively.}
        
\end{figure*}

Following the recent breakthroughs in Arabic Natural Language Processing (NLP), achieved using AraBERT model \cite{antoun2020arabert}, \citet{mageed2020marbert} have introduced two Arabic transformer-based language models, namely ARBERT and MARBERT. ARBERT is trained on large textual corpora of Modern Standar Arabic (MSA), while  MARBERT is trained on 1 billion DA and MSA tweets corpus. They have shown new cutting edge performances on wide range of DA and MSA NLP tasks (AraBench datasets), including, among others, SA and sarcasm detection.                 

In this paper, we present our end-to-end deep MTL model, submitted to SA and sarcasm detection in Arabic language shared task \cite{abufarha-etal-2021-arsarcasm-v2}. Our approach is based on MARBERT \cite{mageed2020marbert}, and a multi-task attention interaction module. The latter consists of two task-specific attention layers for extracting task-discriminative features, and of a Sigmoid interaction layer \cite{LanWWNW17} for allowing interaction and knowledge sharing between sarcasm detection and SA. The task-interaction is performed using the task-specific attention outputs, a learnable shared matrix, and the Sigmoid activation. The obtained results show that our MTL model surpasses the other evaluated single-task and MTL models. Besides, the incorporation of an attention mechanism and the task-interaction boosts the performance of both sarcasm detection and SA.

The rest of the paper is organized as follows. Section \ref{sec:2} presents the shared task's dataset. Section \ref{sec:3} introduces the proposed method. In Section \ref{sec:4}, we present the obtained results for both sarcasm detection and SA subtasks. Section \ref{sec:5} discusses the obtained results. Finally, Section \ref{sec:6} concludes the paper.  

\section{Data}
\label{sec:2}
The ArSarcasm Shared Task consists of two subtasks for sarcasm detection and SA in Arabic language \cite{abufarha-etal-2021-arsarcasm-v2}. The shared task's dataset, ArSarcasm-v2, is built from the previously introduced datasets for sarcasm and irony detection \cite{AbbesZEA20,farha2020arabic}. The provided dataset consists of 12,548 and 3,000 tweets for the training set and test set, respectively. The task's dataset is annotated for  SA and sarcasm detection as well as the regional dialect of the tweets. 

Figure 1 presents the distribution of sarcastic tweets and their sentiment polarities (Figures \ref{fig:sar_reg} and \ref{fig:sen_dist}). The distribution of all sentiment polarities in the dataset is illustrated in \ref{fig:sen_dist}. The dataset is unbalanced for both subtasks. Most sarcastic tweets are written in MSA and Egyptian dialect (Figure \ref{fig:sar_reg}), and are labeled with a negative sentiment (Figure \ref{fig:sar_sen}). Furthermore, approximately half of the tweets convey a neutral sentiment (Figure \ref{fig:sen_dist}).   
\section{Method}
\label{sec:3}

Our multi-task model consists of three main components: BERT encoder, a multi-task attention interaction module, and two task classifiers.

\subsection{BERT Encoder}
Fine-tuning Bidirectional Encoder Representation from Transformers (BERT) model on downstream tasks has shown a new wave of state-of-the-art performances in many NLP applications \cite{devlin-etal-2019-bert}. BERT model's architecture consists of multiple transformer encoders for learning contextualized word embedding of a given input text. It is trained on large textual corpora using two self-supervised objectives, namely the Masked Language Model (MLM) and the Next Sentence Prediction (NSP).  

The encoder of our MTL model is the pre-trained MARBERT \cite{mageed2020marbert}. 
MARBERT is fed with a sequence of wordpeices $[t_1, t_2, ...,t_n]$ of the input tweet, where $n$ is the sequence length. It outputs the tweet embedding $h_{[CLS]}$ ([CLS] token embedding) and the contextualized word embedding of the input tokens $H = [h_1, h_2, ..., h_n] \in \mathbb{R}^{n \times d}$. Both $h_{[CLS]}$ and $h_i$ have the same hidden dimension $d$.

\subsection{Multi-task attention interaction module}
This module consists of two task-specific attention layers (task-specific context-rich representation) and a Sigmoid task-interaction layer. 

The task-specific sentence representation $v_{*} \in \mathbb{R}^{1 \times d}$ (e.g. $v_{sarc}$ and $v_{sent}$) is obtained using the attention mechanism over the contextualized word embedding matrix $H$ :
\[ C = tanh(H W^{a}) \]
\[ \alpha = softmax ( C^{T} W^{\alpha})\]
\[v_{*} = \alpha \cdot H^{T} \]
where $W^{a} \in \mathbb{R}^{d \times 1}$ and $W^{\alpha} \in \mathbb{R}^{n \times n}$ are the learnable parameters of the attention mechanism. $C \in \mathbb{R}^{n \times 1}$ and $\alpha \in [0,1]^{n}$ weights words hidden representations according to their relevance to the task.

The task interaction mechanism \cite{LanWWNW17} is performed using a learnable shared matrix $W^{i} \in  \mathbb{R}^{d \times d}$ and a bias vector $b^{i} \in \mathbb{R}^{d}$. The interaction of both task are given by:
\begin{equation}
    v^{\prime}_{sarc} = v_{sarc} \odot \sigma(W^{i} v_{sent} + b^{i})
\end{equation}
\begin{equation}
    v^{\prime}_{sent} = v_{sent} \odot \sigma(W^{i} v_{sarc} + b^{i})
\end{equation}
where $v_{sarc}$ and $v_{sent}$ are the output of the sarcasm task-specific attention layer and the sentiment task-specific attention layer, respectively. $\odot$ is the element-wise product.
\subsection{Task classifier}
We employ two task classifiers $F_{sarc}$ and $F_{sent}$ for sarcasm detection and SA, respectively. Each classifier consists of one hidden layer and one output layer. They are fed with the concatenation of the pooled output embedding and the task output of the multi-task attention interaction module $v^{\prime}_{*}$ (e.g. $v^{\prime}_{sarc}$ and $v^{\prime}_{sent}$). The outputs of the task classifiers are given by:
\begin{equation}
    \hat{y}_{sarc} = F_{sarc}([h_{[CLS]}, v^{\prime}_{sarc}]) 
\end{equation}
\begin{equation}
    \hat{y}_{sent} = F_{sarc}([h_{[CLS]}, v^{\prime}_{sent}]) 
\end{equation}

\subsection{Multi-task learning objective}
We train our MTL model to jointly minimize the binary cross-entropy loss $\mathcal{L}_{BCE}$, for sarcasm detection, and the cross-entropy loss $\mathcal{L}_{CE}$, for SA. The total loss is given by:
\begin{equation}
    \mathcal{L} = \mathcal{L}_{BCE} (y_{sarc}, \hat{y}_{sarc}) + \mathcal{L}_{CE} (y_{sent}, \hat{y}_{sent})
\end{equation}
where $\hat{y}_{*}$ is the predicted output and $y_{*}$ is the ground truth label.

\section{Results}
\label{sec:4}
In this section, we present the experiment settings and the obtained results.
\subsection{Experiment settings}
We have compared our model (MTL\_ATTINTER) with two single-task models (ST and ST\_ATT) and two MTL models (MTL and MTL\_ATT).
\begin{itemize}
     \setlength\itemsep{0.1em}
    \item \textbf{ST} consists of MARBERT with one classification layer.
    \item \textbf{ST\_ATT}  employs the attention mechanism on top of the contextualized word embedding of MARBERT. The classification is performed using the attention layer output and the [CLS] token embedding.
    \item \textbf{MTL} is similar to \textbf{ST} model and uses classification layer for each task.
    \item \textbf{MTL\_ATT} is the MTL counterpart of \textbf{ST\_ATT} model.
\end{itemize}

\begin{table*}[htbp]
  \centering
  \resizebox{1\textwidth}{!}{%
    \begin{tabular}{llccccc|ccccc}
\cmidrule{3-12}          &       & \multicolumn{5}{c|}{\textbf{Sarcasm}} & \multicolumn{5}{c}{\textbf{Sentiment}} \\
\cmidrule{3-12}          &       & \textbf{Precision} & \textbf{Recall} & \textbf{Accuracy} & \textbf{F1} & \textbf{F$1^{Sarc}$} & \textbf{Precision} & \textbf{Recall} & \textbf{Accuracy} & \textbf{F1} & \textbf{F1$^{PN}$} \\
    \midrule
    \multirow{2}[2]{*}{\textbf{ST}} & \textbf{Dev} & 0.7649 & \emph{0.7683} & 0.8673 & 0.7666 & 0.6132 & 0.7422 & 0.7519 & 0.7641 & 0.7465 & 0.7284 \\
          & \textbf{Test} & 0.706 & 0.708 & 0.768 & 0.707 & 0.573   & 0.672 & 0.667 & 0.713 & \textbf{0.665} & 0.749 \\
    \midrule
    \multirow{2}[2]{*}{\textbf{ST\_ATT}} & \textbf{Dev} & 0.7736 & 0.7588 & 0.8622 & 0.7658 & 0.6156 & \emph{0.7541} & 0.7429 & 0.7629 & \emph{0.7479} & 0.7253 \\
          & \textbf{Test} & 0.724 & \textbf{0.722} & \textbf{0.778} & 0.723 & 0.598   & 0.664 & 0.665 & 0.709 & 0.661 & 0.742 \\
    \midrule
    \multirow{2}[2]{*}{\textbf{MTL}} & \textbf{Dev} & 0.7935 & 0.7611 & 0.8633 & 0.7753 & 0.6347 & 0.7424 & 0.748 & \emph{0.7649} & 0.7448 & 0.7288 \\
          & \textbf{Test} & 0.725 & 0.714 & 0.771 & 0.719 & 0.599  & \textbf{0.676} & 0.656 & 0.703 & 0.662 & 0.736 \\
    \midrule
    \multirow{2}[2]{*}{\textbf{MTL\_ATT}} & \textbf{Dev} & 0.8064 & 0.7581 & 0.8606 & 0.7778 & 0.6421 & 0.7478 & \emph{0.7524} & \emph{0.7649} & 0.7465 & 0.7326 \\
          & \textbf{Test} & \textbf{0.741} & 0.72 & 0.773 & \textbf{0.728} & \textbf{0.617}   & 0.663 & 0.676 & \textbf{0.717} & 0.66 & \textbf{0.752} \\
    \midrule
    \multirow{2}[2]{*}{\textbf{MTL\_ATTINTER}} & \textbf{Dev} & \emph{0.8106} & 0.766 & \emph{0.8661} & \emph{0.7846} & \emph{0.6522} & 0.7511 & 0.7414 & 0.7582 & 0.7436 & \emph{0.7358} \\
          & \textbf{Test} & 0.7268   & 0.7122   & 0.7680   & 0.7183   & 0.6000   & 0.6713   & \textbf{0.7183}   & 0.7107   & 0.6625   & 0.7480 \\
    \bottomrule
    \end{tabular}%
    }%
      \caption{Models evaluation on both SA and sarcasm detection subtasks\label{tab:results}}

\end{table*}%

\begin{table*}[htbp]
  \centering
  \resizebox{1\textwidth}{!}{%
    \begin{tabular}{llccccc|ccccc}
\cmidrule{3-12}          &       & \multicolumn{5}{c|}{\textbf{Sarcasm}} & \multicolumn{5}{c}{\textbf{Sentiment}} \\
\cmidrule{3-12}          &       & \textbf{Precision} & \textbf{Recall} & \textbf{Accuracy} & \textbf{F1} & \textbf{F$1^{Sarc}$} & \textbf{Precision} & \textbf{Recall} & \textbf{Accuracy} & \textbf{F1} & \textbf{F1$^{PN}$} \\
    \midrule
    \textbf{MTL\_ATTINTER}
           & & 0.7268   & 0.7122   & 0.7680   & 0.7183   & 0.6000   & 0.6713   & \textbf{0.7183}   & 0.7107   & 0.6625   & 0.7480 \\
    \bottomrule
    \end{tabular}%
    }%
      \caption{The obtained results of our \textbf{Official} submission \label{tab:offresults}}

\end{table*}%
We have implemented the MARBERT' tweets preprocessing pipeline \cite{mageed2020marbert}. The evaluated models have been trained using Adam optimizer with a learning rate of $5 \times 10^{-6}$. Based on several experiments, the batch size and the number of epochs have been fixed to $64$ and $5$, respectively. Besides, we have used $80$\% and $20$\% of the provided training data for training set and development set, respectively. For comparison purposes, we have used the macro-average Precision, Recall, F1, and F1 score of positive and negative (F1$^{PN}$) evaluation measures. We have also employed the Accuracy and the F1 score of the sarcastic tweets (F1$^{Sarc}$). 
\subsection{Experiment results}
Table \ref{tab:results} shows the obtained models' performances for both SA and sarcasm detection. The best results, for each evaluation measure, are highlighted with italic font and bold fond for the dev set and the test set, respectively. The overall obtained results show that MTL models outperform their single-task counterparts for most evaluation measures. In fact, incorporating attention mechanism into both ST\_ATT and MTL\_ATT improves the F1, F1$^{Sarc}$ and F1$^{PN}$. The former compute the F1 score for sarcastic tweets only, while the latter consider only positive and negative sentiment.

MTL\_ATTINTER and MTL\_ATT achieve the best performances for most evaluation measures on both the dev and the test sets of sarcasm detection sub-task. Specifically, they show far better F1 performance for the sarcastic class prediction.  For SA, the other evaluated models achieve slightly better performance. However, MTL\_ATTINETER and MTL\_ATT yield the best F1$^{PN}$ performances on the dev set and the test set. Therefore, our proposed model excels in detecting sarcastic tweets as well as predicting positive and negative sentiments.  
\subsubsection*{Official results}
Since one submission was allowed, we have submitted the results of our MTL\_ATTINETER model. Table \ref{tab:offresults} shows the official submission results. Our system is top ranked on SA Sub-task and has secured the fourth position among submitted systems for sarcasm detection.
\begin{figure*}[ht]
     \centering
     \begin{subfigure}[b]{0.45\textwidth}
         \centering
         \includegraphics[scale=0.1]{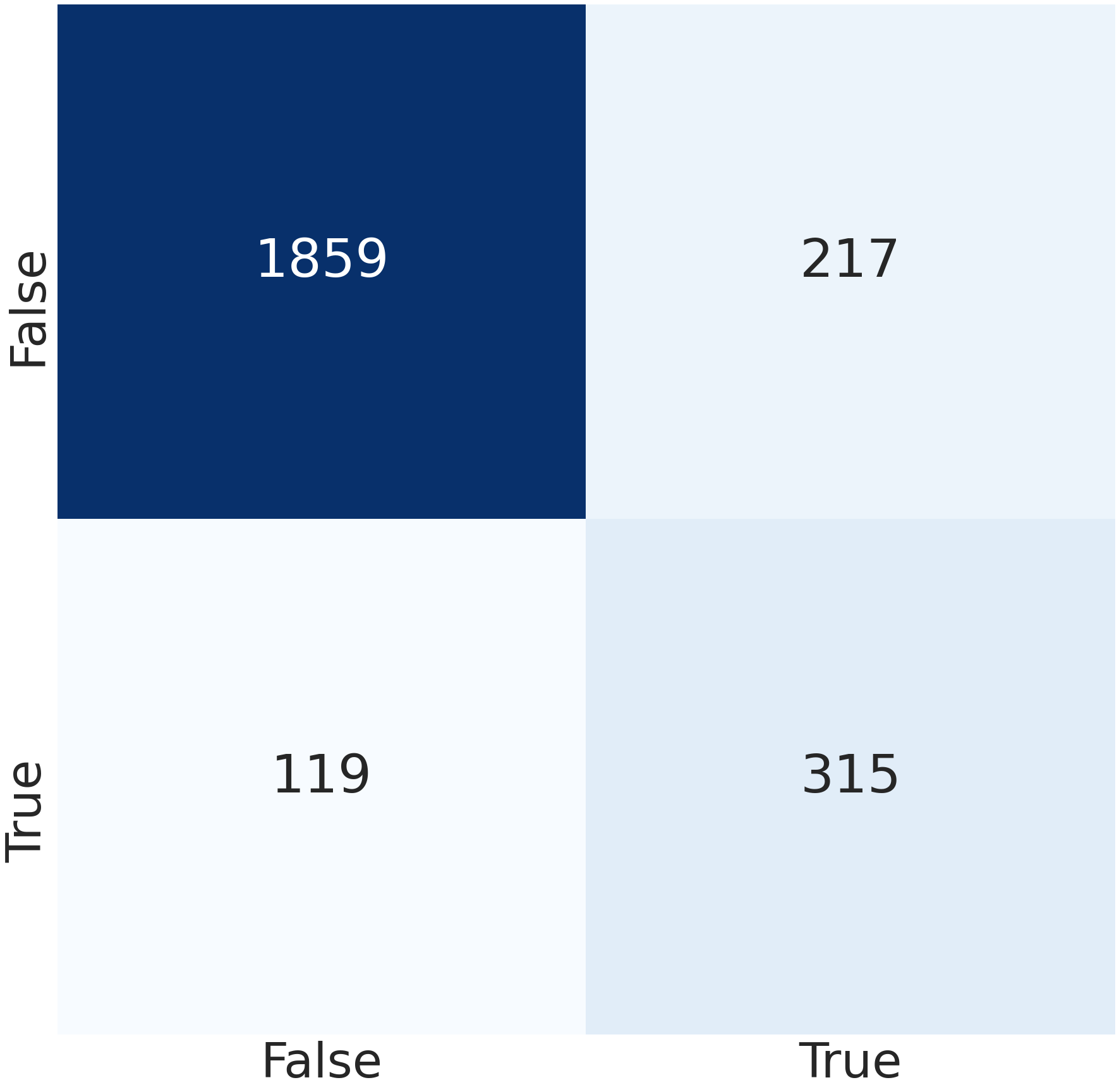}
         \caption{Confusion matrix of the sarcasm detection task}
         \label{fig:sar_cc}
     \end{subfigure}
     \hfill
     \begin{subfigure}[b]{0.5\textwidth}
         \centering
         \includegraphics[scale=0.1]{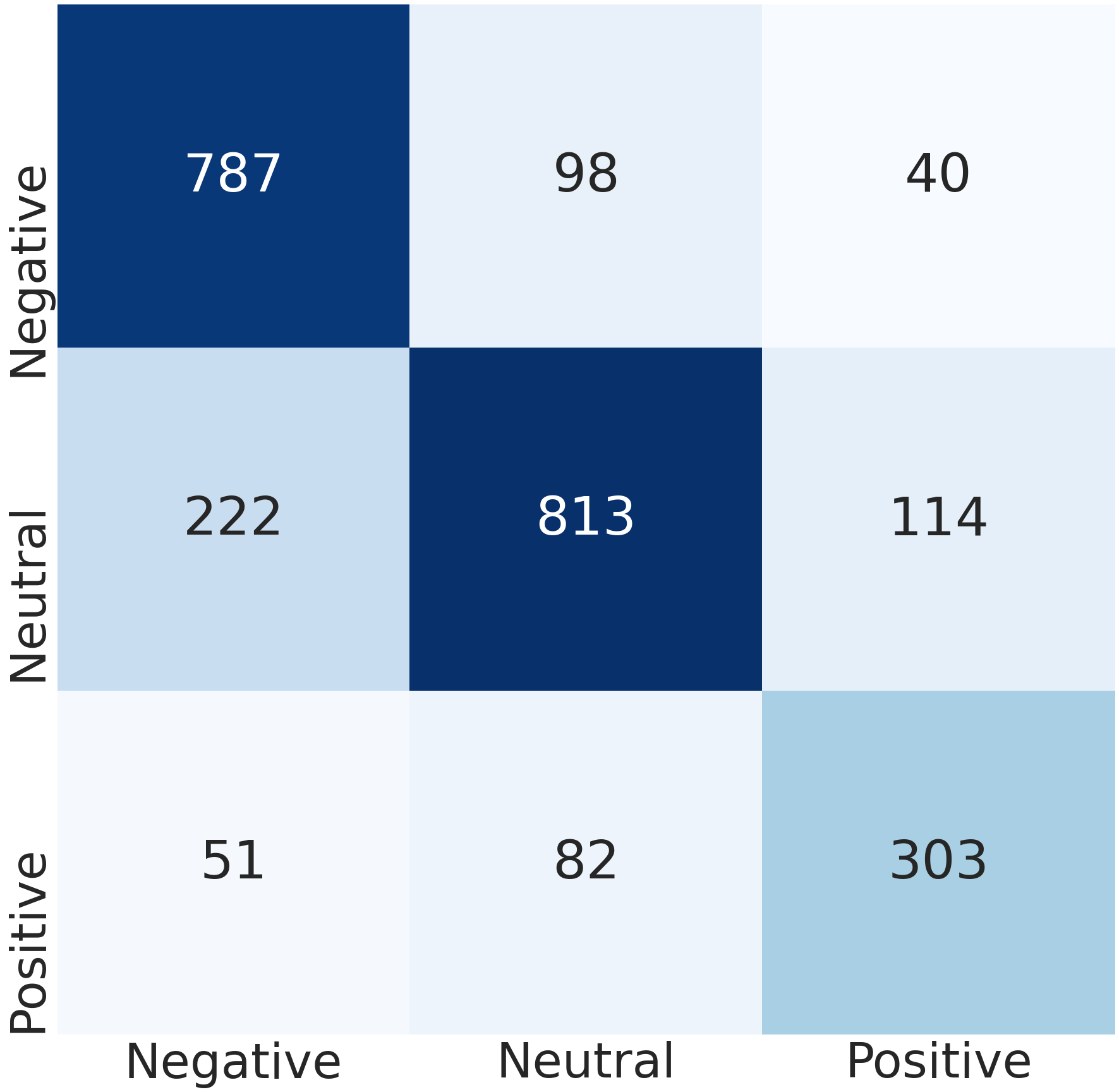}
         \caption{Confusion matrix of SA task}
         \label{fig:sar_sen_cc}
     \end{subfigure}

     \begin{subfigure}[b]{0.45\textwidth}
         \centering
         \includegraphics[scale=0.1]{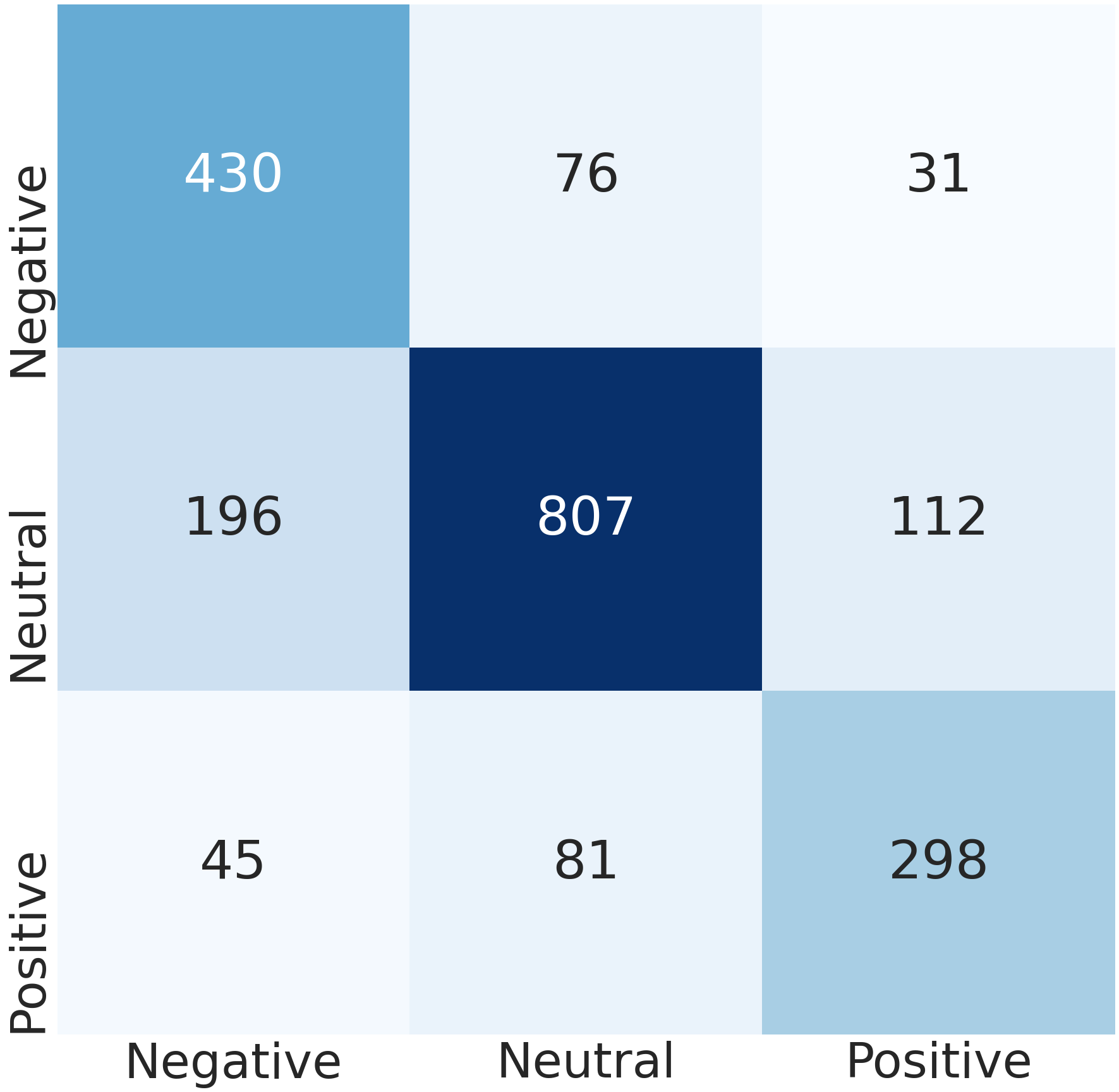}
         \caption{Confusion matrix of SA among non-sarcastic tweets}
         \label{fig:sen_nsarc_cc}
     \end{subfigure}
         \hfill
     \begin{subfigure}[b]{0.45\textwidth}
         \centering
         \includegraphics[scale=0.1]{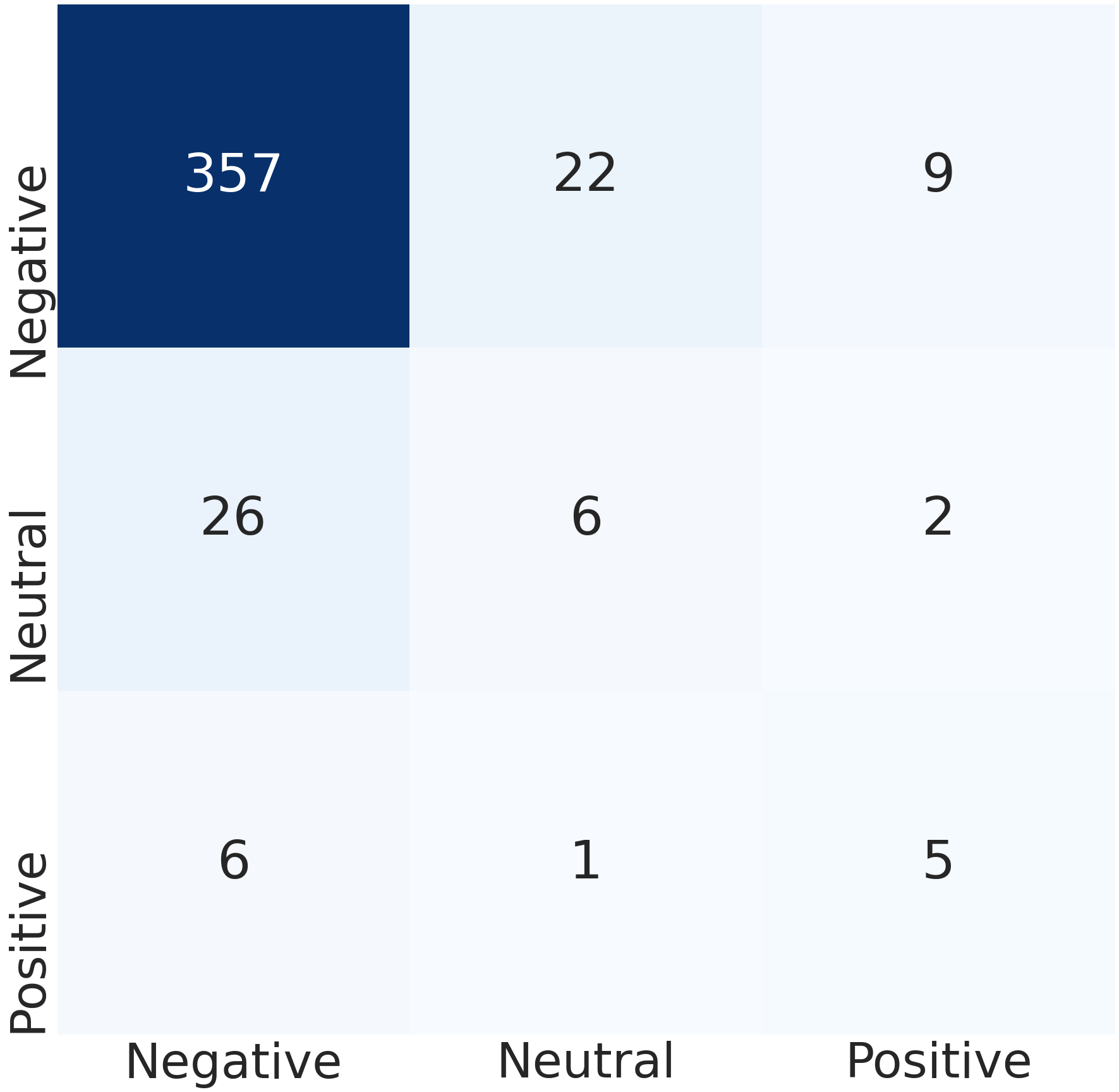}
         \caption{Confusion matrix of SA among sarcastic tweets}
         \label{fig:sen_sarc_cc}
     \end{subfigure}
        \caption{The confusion matrices of our MTL model's prediction on both SA and sarcasm detection tasks}
        \label{fig:arSar}
\end{figure*}
\section{Discussion}
\label{sec:5}
To investigate the strengths and weaknesses of our model, we have analyzed the confusion matrix of each subtask (Figures \ref{fig:sar_cc} and \ref{fig:sar_sen_cc}) as well as the confusion matrices of sentiment analysis among sarcastic and non-sarcastic tweets respectively (Figures \ref{fig:sen_sarc_cc} and \ref{fig:sen_nsarc_cc}). The analysis of these matrices shows that our MTL model leverages signals from both tasks and boosts the performances. This can be explained by the fact that most sarcastic tweets convey a negative sentiment. Besides, negative tweets tend to have a large probability of being sarcastic than the positive ones. This could be also deduced from Table \ref{tab:results}, where MTL models achieve the best F1$^{Sarc}$ and F1$^{PN}$ scores compared to single-task models. 

\section{Conclusion}
\label{sec:6}
In this paper, we have proposed an end-to-end deep Multi-Task Learning model for SA and sarcasm detection. Our model leverages the MARBERT's contextualized word embedding with a multi-task attention interaction module. The aim is to allow task-interaction and knowledge sharing for both SA and sarcasm detection. Our model shows very promising results on both subtasks. Therefore, it proves the effectiveness of using task-specific attention layers as well as the task-interaction mechanism in multi-task learning. 

Future research work will focus on developing task-interaction and class-interaction modules and mechanisms for SA and sarcasm detection.

\bibliography{anthology,eacl2021}
\bibliographystyle{acl_natbib}
\end{document}